\documentclass[reprint,onecolumn,notitlepage,longbibliography]{revtex4-1}
    \usepackage[colorinlistoftodos]{todonotes}  % use if comments are to show in the pdf

    \usepackage{color}      % use if color is used in text

    \usepackage{hyperref}   % use for hypertext links, including those to external documents and URLs 
    \usepackage{color}      % use if color is used in text

    \usepackage{hyperref}   % use for hypertext links, including those to external documents and URLs 

    \usepackage{bm}% bold math
    \usepackage{upgreek} % \upmu
    \usepackage{ulem}
    
    \usepackage{amsmath,amssymb,amsfonts,amsthm,bm,graphicx,epstopdf-base}

    \graphicspath{{./Figures/}}

\begin{document}
    \title{Chemotaxis of sea urchin sperm cells through deep reinforcement learning}
    \author{Chaojie Mo}
    \affiliation{State Key Laboratory of Fluid Power and Mechatronic Systems, Department of Engineering Mechanics, Zhejiang University, Hangzhou 310027, P. R. China}
    \affiliation{Aircraft and Propulsion Laboratory, Ningbo Institute of Technology, Beihang University, Ningbo 315100, P. R. China}
    \author{Xin Bian}
    \email{Corresponding author: bianx@zju.edu.cn}
    \affiliation{State Key Laboratory of Fluid Power and Mechatronic Systems, Department of Engineering Mechanics, Zhejiang University, Hangzhou 310027, P. R. China}
    \date{\today}
\begin{abstract}
 By imitating biological microswimmers, microrobots can be designed to accomplish targeted delivery of cargos and biomedical manipulations at microscale. However, it is still a great challenge to enable microrobots to maneuver in a complex environment. Machine learning algorithms offer a tool to boost mobility and flexibility of a synthetic microswimmer, hence could help us design truly smart microrobots. In this work, we investigate how a model of sea urchin sperm cell can self-learn chemotactic motion in a chemoattractant concentration field. We employ an artificial neural network to act as a decision-making agent and facilitate the sperm cell to discover efficient maneuver strategies through a deep reinforcement learning (DRL) algorithm. Our results show that chemotactic behaviours, very similar to the realistic ones, can be achieved by the DRL utilizing only limited environmental information. In most cases, the DRL algorithm discovers more efficient strategies than the human-devised one. Furthermore, the DRL can even utilize an external disturbance to facilitate the chemotactic motion if the extra flow information is also taken into account by the artificial neural network. Our results provide insights to the chemotactic process of sea urchin sperm cells and also prepare guidance for the intelligent maneuver of microrobots.
\end{abstract}

\maketitle

\section{Introduction}
Biological microswimmers live in the regime of low Reynolds number~\cite{Purcell1977},
where the viscous force dominates the inertial force at microscale. As a result, they propel themselves using strategies completely different from that of macroscopic organisms. Evolution has led the microscopic organisms to develop effective propellers such as wriggling flagellum and rotating helix~~\cite{Berg1973}, which can overcome and even exploit the overwhelming viscous forces~\cite{Lauga2009,Elgeti2015}. In particular, bacterial flagellum is one of the most renowned propeller at microscale~\cite{Fauci2005,Liu2007}.
To understand life at microscale, it is crucial to investigate not only the biological structures of the self-propelling organism, but also their propulsion mechanism from the perspective of fluid dynamics~\cite{Happle1983,Kim1991}. It is for this reason that the microscopic propulsion has drawn lots of attentions and fruitful results have been reported in the past decades~\cite{Lauga2009,Elgeti2015}. Recently, research focus has been further shifted to the propulsion mechanism in complex fluids and environments~\cite{Riley2015,Arratia2017,Tung2017,Ishimoto2018}.

Studies on microswimmers not only help us understand the nature but may also  teach us to design synthetic microswimmers. These microrobots may be used for cargo deliveries and biomedical manipulations in microfluidic and even {\it in vivo} systems, hence have a great potential for non-invasive drug delivery and medical treatment. Many synthetic microswimmers have been invented and further successfully applied: catalytic Janus particles exploiting the diffusiophoresis or thermophoresis process to accomplish self-propulsion~\cite{Paxton2004,Howse2007,Jiang2010,Volpe2011}; catalytic nanomotors propelled by generating bubble jet~\cite{Sanchez2011};
self-propelling droplet by the Marangoni stress in a surfactant solution~\cite{Li2022};
rotators breaking kinetic symmetry and actuated by external magnetic field~\cite{Tierno2008}; biohybrid microswimmers imitating sperm cells~\cite{Dreyfus2005,Sanchez2011Science,Williams2014}; and neutrophil-based microrobots that can actively deliver cargo to malignant glioma through chemotactic motion~\cite{Zhang2021}. These examples evidenced microswimmers as an promising research direction due to the great potential.
However, if the drug delivery is to be accomplished, there are at least two issues to resolve: the microswimmer must survive the immune attack and be able to cross biological barriers; it must maneuver precisely to the target through a complex environment. In this work, we address ourselves to the maneuvering problem.

There are many ways to steer a microswimmer along a specific direction, such as chemotaxis~\cite{Saha2014,Jin2017}, magnetotaxis~\cite{Ahmed2017,Baraban2012,Sanchez2011,Amoudruz2022}, phototaxis~\cite{Dai2016,Lozano2016}, gravitaxis~\cite{Campbell2013,TenHagen2014}, viscotaxis~\cite{Liebchen2018,Datt2019} and so on. Among these, the magnetotaxis is one of the most frequently adopted in laboratory due to its non-invasive characteristics and high efficiency on maneuvering. But the disadvantage of magnetotaxis is also apparent: it needs a large system to generate the rotational magnetic field, and can usually manipulate only one microswimmer at a time. In contrast, biological microswimmers, such as sperm cells~\cite{Friedrich2007}, E. coli~\cite{BergHoward1972} and green algae~\cite{Choi2016}, often follow the chemotactic process to swim by themselves towards the target or to seek food. Biological microswimmmers have learnt chemotaxis through millions of generations in evolution, which also inspired researchers to design synthetic microswimmers to accomplish chemotaxis in specific environments~\cite{Saha2014,Jin2017}. 
Nevertheless, what remain largely open is: how can a synthetic microswimmer be programmed to learn chemotaxis by itself, hence adjust itself automatically to different complex environments? 
Answering this question would significantly improve the navigation ability of a synthetic microswimmer, therefore lead to  an ``intelligent'' microrobot rather than a nonrational one.

Recently, there have been already some efforts to design intelligent microswimmers using machine learning techniques, especially the reinforcement learning (RL) algorithms. Colabrese \textit{et al}~\cite{Colabrese2017} and Gustavsson \textit{et al}~\cite{Gustavsson2017} applied the Q-learning method to train active gravitatic microswimmers to accomplish counter-gravity navigation through 2D Taylor-Green vortex flow and 3D chaotic flow field. Tsang \textit{et al}~\cite{Tsang2020} also adopted Q-learning to train the Najafi-Golestanian swimmer to self-learn propulsion. They found that when the structure of the swimmer becomes complex, the RL algorithm can discover propulsion mechanism much more efficient than a human-designed one. Alageshan \textit{et al}~\cite{Alageshan2019} studied the path-planning problem through a complex turbulent flow field and they employed a multiswimmer adversarial Q-learning algorithm to find the optimised steering strategy towards a specified target. They found that if the adversarial Q-learning scheme is applied, the average time required to reach the target is smaller than that of a simple strategy. Gunnarson \textit{et al}~\cite{Gunnarson2021} also studied the navigation of a swimmer in a time-varying vortical flow field, where they feed the environmental cues to a deep neural network that determines the swimmer's action, and train the network with a experience replay. As a result, the swimmers successfully discovered efficient policies to reach the target. Yang \textit{et al}~\cite{Yang2020} kept a Janus particle to rotate randomly to perceive the obstacles around itself, and thereafter applied a deep reinforcement learning (DRL) algorithm to train the Janus particle. They showed that the Janus particle guided by the deep convolutional Q-network can act smartly to bypass the obstacles and swim toward its target.  Hartl \textit{et al}~\cite{Hartl2021} studied the self-learned chemotaxis of a Najafi-Golestanian swimmer in 1D space, where they decoupled the task into two parts. One is to teach the swimmer to swim, and the other is to train the swimmer to determine the gradient direction of the chemoattractant concentration field and steer itself to that direction. They applied the neural evolution of augmenting topologies (NEAT) to optimize not just the weights of the neural network but also the topology. They found very simple architectures of the neural network to accomplish the chemotaxis task. These simple neural networks provide insights on how simple biological microswimmers are able to sense the environment and achieve chemotactic motion, which may be highly realizable on synthetic microswimmers. In summary, the RL algorithms have been proven to be a very powerful tool for intelligent control of swimmers\cite{Amoudruz2022,zhu_fang_zhu_2022,Verma2018}.

Despite the successes of applications of RL listed above, we have to mention that most of the work either completely neglected the microswimmer's structural details, that is, they are assumed to be active material points, or employed very simple ones such as Janus particle or three-beads swimmer. Furthermore, they often ignored the detailed steering mechanism except in the models of Janus particle and three-beads. Usually no comparison to the chemotaxis of biological microswimmers has been made. It should bear in mind that the chemotactic motion depends strongly on the steering mechanism. For example, the E.\ coli empolys the run-and-tumble strategy to achieve the chemotactic motion, since it can only change its swimming direction randomly through unbundling its rotating helical flagella occasionally~\cite{Berg1973}; while a sea urchin sperm cell achieves the chemotactic motion by bending its helical path towards the gradient direction, as it can steer by altering the waveform of propagating wave on its flagellum~\cite{Friedrich2007}. Therefore, it is imperative to consider the detailed steering mechanism when studying the self-learned chemotaxis of microswimmers. In this work, we shall take the sea urchin sperm cell as model and try to imitate its steering mechanism by applying the DRL to learn the chemotactic motion.

We show that with little environmental information the sperm cell can self-learn to achieve a chemotactic motion similar to a realistic one. In most cases, the strategies discovered by the DRL is more efficient than a human-devised one. By loading the external flow information to the neural network, the sperm cell can also learn to exploit the external flow to facilitate its chemotactic motion, hence actively adjusting itself to the complex environment. Our results provide insights to the chemotaxis process of sea urchin sperm cells and also prepare guidance for the smart maneuver of synthetic sperm-like microswimmers.

\section{Model description}
\subsection{Microswimmer model}
\label{sec:microswimmer_model}
Sperm cells swim by generating propagating waves of deformation along their flagella and steer through non-zero average curvatures~\cite{Eshel1987,Friedrich2007} and/or second harmonics~\cite{Saggiorato2017}. 
Although it has been shown that the first steering mechanism has higher effectiveness~\cite{Gong2020},  there is a similar transfering process of signal in both mechanisms.
In short, the chemoattractants in the environment bind the receptors on the surface of the sperm cells and initiate a signalling cascade to change the intracellular concentration of $\text{Ca}^{2+}$. Moreover, the $\text{Ca}^{2+}$ concentration influences the activity of the dynein motors, which further modulate the beating waveform of the flagella. The sperm cells rely on the whole process above to steer. Phenomenologically, the chemotactic signalling network initiates a series of complex tasks: temporally comparing the chemoattractant concentrations along the swimming path; determining the gradient direction of the concentration field; and steering the sperm to swim toward the gradient direction upwards. In the works of  Friedrich \textit{et al}~\cite{Friedrich2007,Friedrich2009,Kromer2018}, an adaptive dynamical system has been designed to imitate the behaviour of the chemotactic signalling network and capture the essence of the chemotaxis behaviour. However, in this work, we intend to allow the sperm cells to accomplish self-learned chemotaxis. Therefore, an artificial neural network is employed to replace the chemotactic signalling network inside the sperm cells and to perform the task of decision-making. Before going into the details of the artificial neural network, we first demonstrate the model of swimming path. 

The swimming model from Friedrich \textit{et al}~\cite{Friedrich2007,Friedrich2009,Kromer2018} is employed here. The interacting details between the flagellum and the fluid are neglected and the position of a sperm cell is represented by the average position of the head in one beating cycle. The swimming path is then described by the  Frenet-Serret equation in 2D:
\begin{equation}
	\dot{\mathbf{r}} = v \mathbf{t}, \quad \dot{\mathbf{t}} = v\kappa \mathbf{n}, \quad \dot{\mathbf{n}} = -v\kappa \mathbf{t},
\end{equation}
where $\mathbf{r}$ is the position vector of the sperm cell, $\mathbf{t}$ and $\mathbf{n}$ are the unit tangential vector and unit normal vector, respectively. $v$ is the swimming speed, $\kappa$ is the curvature of the swimming path that will be modulated by an agent. 
The reason for the employment of this model is twofold: firstly, its dynamics is sufficiently complex,
which imitates the realistic behaviors of sperm cells; secondly, it does not involve difficult computations
of fluid-structure interactions so that we can devote more efforts to discover the intelligent steering strategies.

\subsection{Deep Q-Network as decision-making agent}
\begin{figure}[hbt]
\centering
  \includegraphics[width=0.5\linewidth]{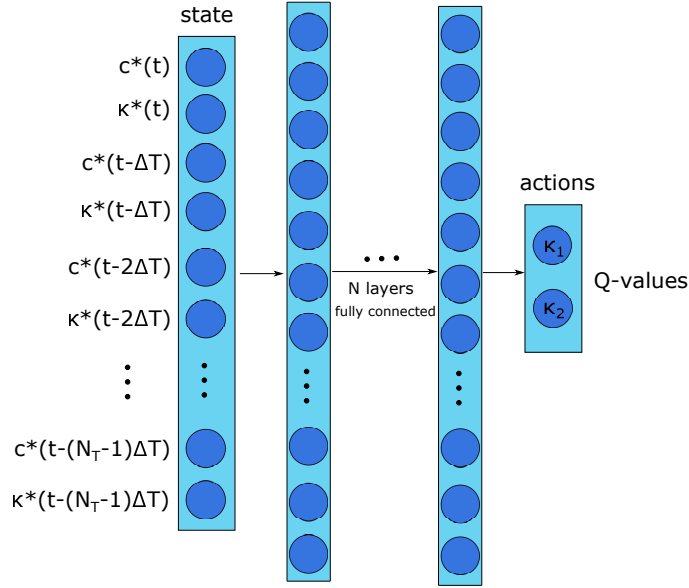}
  \caption{Schematic of the Deep Q-Network. The inputs nodes represent the states of the swimmer including current and historical states. The output nodes represent the legal actions, their values are their corresponding Q-values.}
\label{fig:nn_schematic}
\end{figure}

We consider sperm cells that swim in a field of chemoattractant,
which is described by the concentration $c(\mathbf{r})$. Each sperm cell is aware of the concentration in its current position and can also remember it for a period of time. 
In the context of reinforcement learning, the concentration information together with the cell's curvatures form the state $s$. A so-called agent can take in the state information and propose the next action $a$ for the sperm cell to maximise a reward function. The proposed action is then utilized to update the dynamics of the sperm cell. As we employ an artificial neural network (Deep Q-Network) to represent the agent, as shown in Fig.~\ref{fig:nn_schematic}, the input vector consists of a temporal sequence of the concentrations and path curvatures: 
	\begin{equation}
	\left\{
		\begin{array}{c}
		c(t),\kappa(t), \\
		c(t-\Delta T),\kappa(t-\Delta T), \\
		\vdots \\
		c(t-(N_T-1)\Delta T),\kappa(t-(N_T-1)\Delta T),
		\end{array}
	\right.
	\label{eq:simple_input}
	\end{equation}
where $N_T$ is the number of discrete time points to record the concentrations and curvatures.
Therefore, a state $s$ has $2N_T$ elements.
The output nodes represents the legal actions, and their values are the Q-values. The input vector is connected to the output neurons through several layers of hidden neurons. All the adjoining layers are fully connected. The tanh activation function is used for the input and hidden layers while the linear activation function is used for the output layer. A deep Q-learning procedure~\cite{Mnih2013} is applied to update the weights of the connections $\theta_i$ aiming to approximate the maximum action-value function $Q^*(s,a)$ via the Deep Q-Network. The reward is represented as $r(s,a,s')$ when the sperm cell takes the action $a$ and the state transfers from $s$ to $s'$. The network should predict a target Q-value:
    \begin{equation}
	  Q_\mathrm{target} = r(s,a,s') + \gamma \mathrm{max}_a Q(s',a;\theta_{i}).
    \end{equation}
    where $\gamma$ is the discount-rate parameter. The Q-value predicted by the Deep Q-Network is $Q_\mathrm{predicted}(s,a,\theta_i)$. Therefore the loss function can be defined as:
    \begin{equation}
    	L_i(\theta_i) = E_\pi \left[Q_\mathrm{target}(s,a;\theta_{i}) - Q_\mathrm{predicted}(s,a;\theta_i)\right]^2,
    \label{eq:loss}
    \end{equation}
which is to be minimized.
Note that the same Q-Network with parameter $\theta_i$ is employed for both $Q_\mathrm{target}$ and $Q_\mathrm{predicted}$.
The experience replay technique~\cite{Mnih2013} is applied to randomize the experience data and alleviate the autocorrelation problem. Therefore, a replay buffer of finite size is created at the beginning of the learning. Every $\Delta T$ (which will be rigorously defined in Eq.~(\ref{eq:Delta_T})) the agent's experience $e_t=(s,a,r,s')$ is stored into this buffer. During learning, a minibatch of experience is randomly drawn from the buffer to update the Q-Network using Eq.~(\ref{eq:loss}). The experience is replayed every time as the simulated swimmer reaches the end of its lifespan ($t_\mathrm{life}$). The finish of a replay marks the end of an episode of the learning. Afterwards, a new swimmer with random initial position and direction is created to start a new episode.
    
 During the simulation of each swimmer, the $\epsilon$-greedy policy is used by the agent to select its actions, which balances the exploration and exploitation. An action is selected according to the probability:
   \begin{equation}
   	\pi(a|s) = \begin{cases}
   		\epsilon/m+1-\epsilon, & \text{if}\quad a^*=\mathrm{argmax}_a Q(s,a) \\
   		\epsilon/m, & \text{otherwise}
   	\end{cases}
   \end{equation}
where $m$ is the number of legal actions, $\pi(a|s)$ is the probability to select action $a$ at state $s$. Note that the first row is the probability for the optimal action to be selected, the second row is the probability of other actions to be selected. Since the $\epsilon$-greedy policy is applied, we first generate a random number from the uniform distribution. If the random number is smaller than $\epsilon$, we pick an action randomly among all actions with equal probabilities:
the probability for the optimal action to be selected is $\epsilon/m$; If the random number is larger than $\epsilon$, the optimal action is certainly selected, where the probability is $1-\epsilon$. Therefore, the total probability for the optimal action to be selected is $\epsilon/m + 1-\epsilon$. For every other action, the probability is $\epsilon/m$
and there are $m-1$ ones.
As a matter of fact, if we multiply $m-1$ to the probability of the second row and add to the first row, we get $1$.
   
For simplicity, we consider only two legal actions here ($m=2$), that is,  two values of the path curvature:  $a_t \in \{\kappa_1, \kappa_2\}$ with $0< \kappa_1 < \kappa_2$. The value of $\epsilon$ starts with $1.0$ and slowly anneals to $0.1$ during the training.

Note that the values of input vector are normalized before they are actually taken to the network. Assuming that there are $N_T$ records of concentration information and the time interval for the records is $\Delta T$, the concentration values are normalized as:
\begin{equation}
	c^*(t-i\Delta T) = \left[c(t-i\Delta T) - \frac{1}{N_T}\sum_{j=0}^{N_T-1} c(t-j\Delta T)\right]\frac{\bar{\kappa}}{c_k},\quad i=0,1,\cdots,N_T-1
\end{equation}
where $\bar{\kappa}$ is the characteristic path curvature $\bar{\kappa}=(\kappa_1+\kappa_2)/2$ and $c_k$ is the typical gradient of the concentration field. The $N_T$ records of the curvature information are also normalized: 
\begin{equation}
	\kappa^*(t-i\Delta T) = 2\frac{\kappa(t-i\Delta T)-\bar{\kappa}}{|\kappa_1-\kappa_2|},\quad i=0,1,\cdots,N_T-1.
\end{equation}

In the realistic chemotactic process, a sea urchin sperm cell swims in a helical path to sample the concentration field. It perceives periodic stimulus from the environment and regulates the curvature and torsion of the path accordingly so that it can bend the path toward the chemoattractant source. In light of this, we need to allow our decision-making agent to remember the past concentration information at least for one average period $T$. Here one average period is defined as 
\begin{equation}
T=\frac{2\pi}{\bar{\kappa}v}.
\label{eq:T}
\end{equation}
In this work we focus chemotaxis in 2D. Therefore, the sperm cells swim in drifting circular path. It is reasonable to set $N_T\Delta T \approx T$ so that the information of a fragmental trajectory in the past, which is equivalent to a whole circle in time, is available for the sperm cell. Then $N_T$ is just the number of samples on the fragmental trajectory. A larger $N_T$ indicates a finer perception and probably a more accurate chemotaxis, but may also require larger network and causes the training time to increase dramatically. Here we consider only $N_T=2,4,8$.  We specify $N_T$ first and then set 
\begin{equation}
\Delta T = \mathrm{int}\left(\frac{T}{N_T \Delta t}\right)\Delta t,	
\label{eq:Delta_T}
\end{equation}
with $\Delta t$ being the integration time step of the swimmer's dynamics. 
The interval $\Delta T$ is also the action interval. Assuming that the state of the sperm cell is $s$ at time $t$, the agent takes the action $a$ so that the state of the sperm cell transfers to $s'$ at time $t+\Delta T$, the reward function is defined as:
\begin{equation}
	r(s,a,s') = \frac{1}{c_k|1/\kappa_1-1/\kappa_2|}\left[\frac{1}{N_T}\sum_{j=-1}^{N_T-2} c(t-j\Delta T) - \frac{1}{N_T}\sum_{j=0}^{N_T-1} c(t-j\Delta T)\right] = \frac{c(t+\Delta T) - c(t-(N_T-1)\Delta T)}{N_T c_k|1/\kappa_1-1/\kappa_2|}.
\end{equation}

We train the sperm cell in a linear concentration field:
\begin{equation}
	c = c_k^l y + c_0^l,
	\label{eq:linear_conc_field}
\end{equation}
and test the trained Q-network mainly in the same linear concentration field. But the radial concentration field is also considered:
\begin{equation}
	c = c_0^r-c_k^r\sqrt{x^2+y^2}.
	\label{eq:radial_conc_field}
\end{equation}

Note that every time the sperm cell changes its path curvature $\kappa_1\rightarrow \kappa_2$, there is a displacement on the curvature center: 

\begin{equation}
	\Delta \mathbf{r}_c = -\mathbf{n}(\frac{1}{\kappa_1}-\frac{1}{\kappa_2}).
\end{equation}
We can devise a naïve strategy that guides the cell to swim toward higher concentration:
\begin{equation}
    \kappa: 
	\begin{cases}
		\kappa_2, & \text{if}\quad c(t)=\mathrm{max}(c(t),c(t-\Delta T),\cdots,c(t-(N_T-1)\Delta T)) \\
		\kappa_1, & \text{if}\quad c(t)=\mathrm{min}(c(t),c(t-\Delta T),\cdots,c(t-(N_T-1)\Delta T)) \\
		\text{stay unchanged}, & \text{otherwise}.
	\end{cases}
\end{equation}
We call this the greedy strategy, and shall compare our DRL results with it.
\subsection{External perturbation}
The case in the presence of an external flow field is also considered,
where the 2D Taylor-Green (TG) vortex is adopted:
\begin{equation}
\begin{gathered}
\mathbf{u}_e = [u,v]\\
u_x=u_0\cos(kx)\sin(ky)\\
u_y=-u_0\sin(kx)\cos(ky).
\end{gathered}
\label{eq:taylor_green}
\end{equation}
In this case, the equations of motion for the sperm cell becomes:
\begin{equation}
	\begin{gathered}
	\dot{\mathbf{r}} = v\mathbf{t} + \mathbf{u}_e \\
	\dot{\mathbf{t}} = (v\kappa + \omega_0)\mathbf{n} \\
	\dot{\mathbf{n}} = -(v\kappa + \omega_0)\mathbf{t},
	\end{gathered}
\end{equation}
where $\omega_0=-u_0k\cos(kx)\cos(ky)$ is the angular velocity determined from the vorticity field. The sperm cell is also aware of its external flow field information and we investigate whether the sperm cell can utilize this information to facilitate its chemotaxis. For this purpose the Q-Network must take into account the external flow field information, 
hence the input vector contains $5N_T$ elements:  
\begin{equation}
	\left\{ 
	\begin{array}{c}
	c(t),\kappa(t),u_x(t), u_y(t), \omega_0(t), \\
	c(t-\Delta T),\kappa(t-\Delta T),u_x(t-\Delta T), u_y(t-\Delta T), \omega_0(t-\Delta T), \\
	\vdots \\ 
    c(t-(N_T-1)\Delta T),\kappa(t-(N_T-1)\Delta T), u_x(t-(N_T-1)\Delta T), u_y(t-(N_T-1)\Delta T), \omega_0(t-(N_T-1)\Delta T).
	\end{array}
	\right.	
	\label{eq:rich_input}
\end{equation}
The velocity $u_x$ and $u_y$, and the angular velocity $\omega_0$ are normalized before they are inputed into the Q-network:
\begin{equation}
	\begin{gathered}
	u_x^* = u_x/u_0\\
	u_y^* = u_y/u_0\\
	\omega_0^* = \omega/(u_0k).
	\end{gathered}
\end{equation}
For simplicity we consider a weak disturbance flow ($u_0/v=0.1$) and assume that the spatial distribution of the chemoattractant is not affected by the flow field. 
\subsection{Simulation parameters}
The simulation parameters are summarized in Table \ref{tab:model_parameters}
while the DRL training parameters are summarized in Table \ref{tab:DRL_parameters}. 
If not stated otherwise, values in the two tables are adopted.  
All simulations and trainings run with Python 3.7 and Tenorflow 2.6.0 on the Windows 10 desktop installed on Intel(R) Core(TM) i9-11900K CPU.
\begin{table*}
\centering
\caption{Basic parameters for the simulations}
\label{tab:model_parameters}
\begin{tabular}{lc}
  \textbf{Parameters} & \textbf{Values}\\
\hline  
Swimming velocity $v$ &  1.0 \\
Minimum path curvature $\kappa_1$  & 3.0 \\
Maximum path curvature $\kappa_2$ & 5.0 \\
Minimum swimming velocity $v_2$ & 0.9 \\
Maximum swimming velocity $v_1$ & 1.1 \\
Integration time step $\Delta t$ & 0.02 \\
Physical integration time in training $t_\mathrm{life}$ & 80 \\
Physical integration time in tests $t_\mathrm{life}$ & 200 or 400 \\
Concentration gradient parameter $c_k^l$ & 1 \\
Concentration constant $c_0^l$ & 20 \\
Concentration gradient parameter $c_k^r$ & 1 \\
Concentration constant $c_0^r$ & 100 \\
TG flow field parameter $u_0$ & 0.1 \\
TG flow field parameter $k$ & $\pi/10$ \\

\hline
\end{tabular}
\end{table*}

\begin{table*}
\centering
\caption{Basic parameters for the DRL trainings}
\label{tab:DRL_parameters}
\begin{tabular}{lc}
  \textbf{Parameters} & \textbf{Values}\\
\hline
Learning rate $\alpha$ & 0.01 \\
Learning rate decay & 0.1 \\
Discount-rate parameter $\gamma$ & 0.98 \\
$\epsilon$-greedy parameter $\epsilon$ & 0.1 \\
Number of hidden dense layers $N_\mathrm{hidden}$ & 3 \\
\hline 
\multicolumn{2}{c}{Trainings without TG} \\
\hline
Number of nodes in each hidden layer $N_n$ & 24\\
Epochs & 1600 \\
$\epsilon$ decaying rate & 0.998 \\
\hline 
\multicolumn{2}{c}{Trainings with TG} \\
\hline
Number of nodes in each hidden layer $N_n$ & 36\\
Epochs & 6000 \\
$\epsilon$ decaying rate & 0.9996 \\
\hline
\end{tabular}
\end{table*}

\section{Results and discussion}
\subsection{Perception and action frequency}
\begin{figure}[hbt]
\centering
  \includegraphics[width=0.5\linewidth]{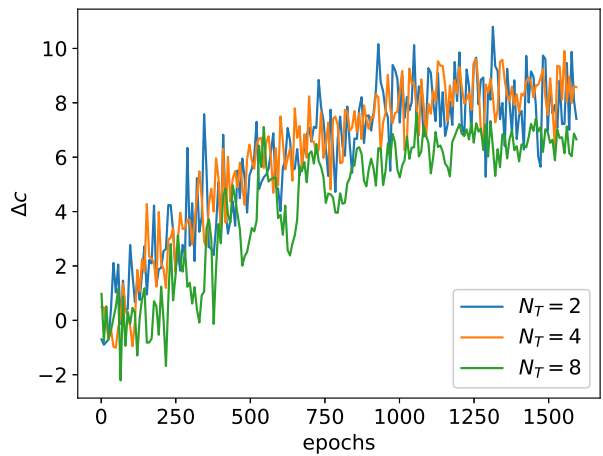}
  \caption{The evolution of the gain $\Delta c$ during trainings.}
\label{fig:training}
\end{figure}

As already explained above, $N_T$ denotes how many records of perception are used as input information for the decision-making machinery. We set $N_T \in \{2,4,8\}$ and perform training for 1600 epochs in each case. During the training processes the gain of the sperm cell $\Delta c$, which is defined as the chemoattractant concentration the cell perceives at the end of each swimming simulation minus the initial concentration $\Delta c = c(t_\mathrm{life}) - c(0)$, increases with the epoch as shown in Fig.~\ref{fig:training}.

\begin{figure}[hbt]
\centering
  \includegraphics[width=0.9\linewidth]{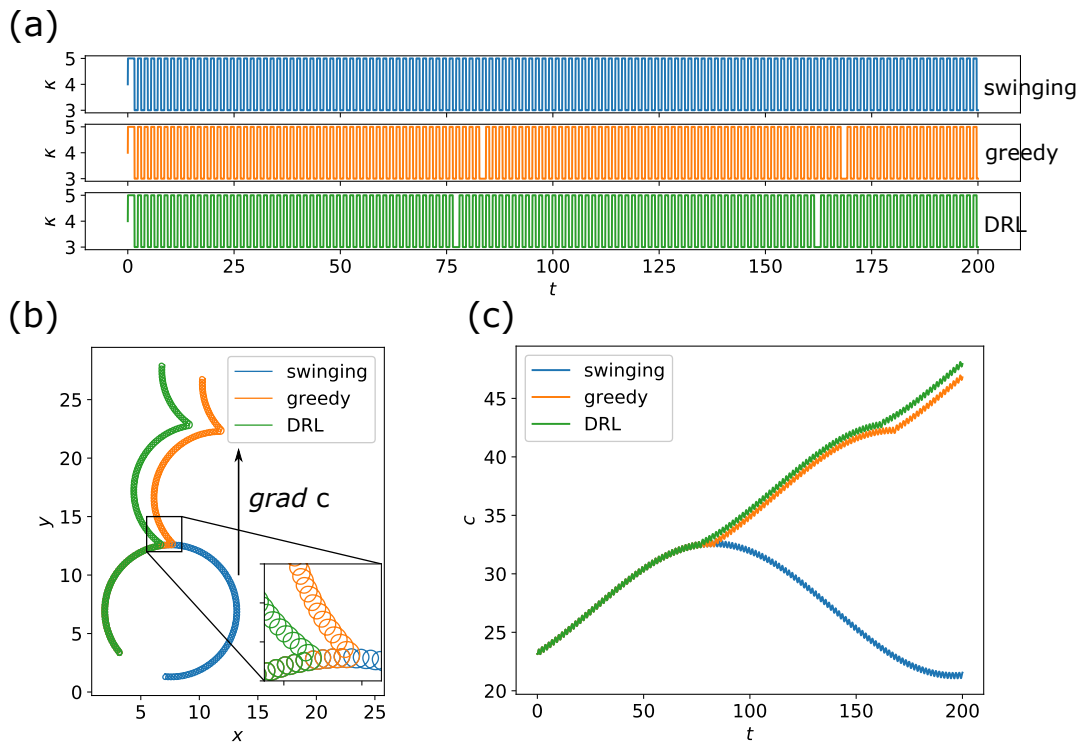}
  \caption{Comparison of (a) evolution of $\kappa$, (b) swimming trajectories, and (c) gains of chemoattractant for a sperm cell using different steering strategies. $N_T=2$, $t_\mathrm{life}=200$.}
\label{fig:sec2_comp}
\end{figure}

\begin{figure}[hbt]
\centering
  \includegraphics[width=0.9\linewidth]{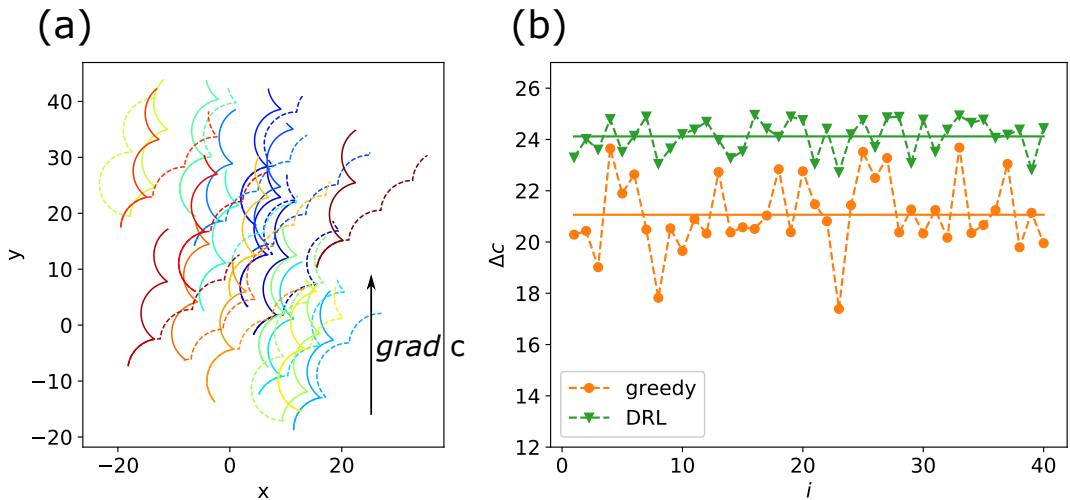}
  \caption{(a) Centerlines of some test swimming sperm cells: solid lines for the DRL; dash lines for the greedy strategy. (b) The final gains ($\Delta c=c(t_\mathrm{life})-c(0)$) of 40 sperm cells. The solid lines mark the average values, respectively. $N_T=2$, $t_\mathrm{life}=200$.}
\label{fig:sec2_trjs}
\end{figure}

We first examine the results of $N_T = 2$. In Fig.~\ref{fig:sec2_comp}, we show how a sperm cell trained with DRL adjusts its curvature and swims toward higher concentration of chemoattractant. For comparison, we also show the results using the periodically swinging curvature pattern and the greedy strategy. In the swinging pattern, the curvature changes every $N_T \Delta T/2$. As shown in Fig.~\ref{fig:sec2_comp} (a),  most of the time, both the DRL and the greedy strategy follow closely the swinging pattern. Fig.~\ref{fig:sec2_comp} (b) shows that the sperm cell swim in shifted circles and the centerline of the path is an arc when the swinging pattern is adopted. However, the sperm cell would swim back to its initial position if the pattern were not broken at some point. Both the greedy strategy and DRL break the swinging pattern, that is, switch between $\kappa_1$ and $\kappa_2$, when the centerline of the path starts to go toward  lower concentration.  In general, the DRL generates a very similar strategy to the greedy one to steer the sperm cell toward a higher concentration,
whereas the timing to break the pattern is different for the two strategies. 
Moreover, Figure \ref{fig:sec2_trjs} (a) shows the centerlines of many test sperm cells, while Figure \ref{fig:sec2_trjs} (b) shows their final gains. 
Although the overall swimming directions of the sperm cells guided by the DRL and the greedy strategy are not perfectly in line with the gradient direction of the chemoattractant,
the DRL achieves visually better alignment of path with the gradient direction,
as shown in Figure \ref{fig:sec2_trjs} (a).
Therefore, the average final gains of the DRL is also always higher than that of the greedy strategy. This indicates that the DRL algorithm has discovered a better timing to break the swinging curvature pattern.

\begin{figure}[hbt]
\centering
  \includegraphics[width=0.9\linewidth]{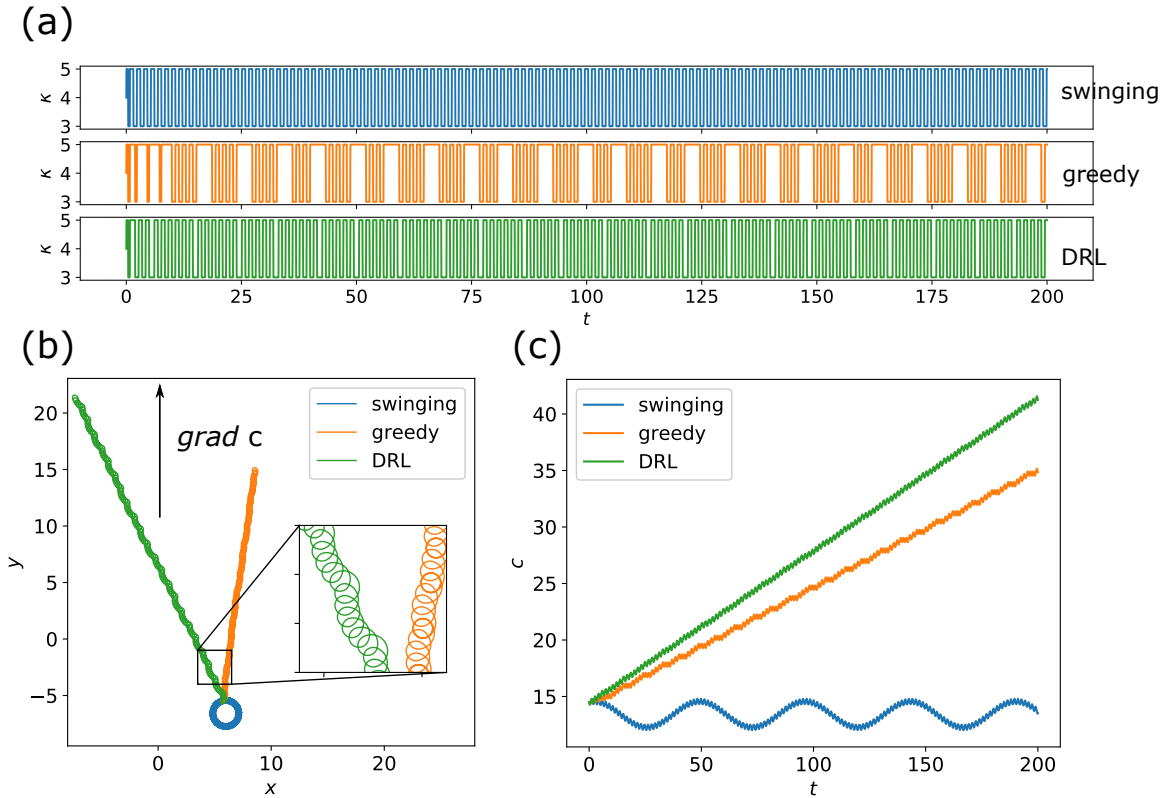}
  \caption{Comparison of (a) evolution of $\kappa$, (b) swimming trajectories, and (c) gains of chemoattractant for a sperm cell using different steering strategies. $N_T=4$, $t_\mathrm{life}=200$.}
\label{fig:sec4_comp}
\end{figure}

\begin{figure}[hbt]
\centering
  \includegraphics[width=0.9\linewidth]{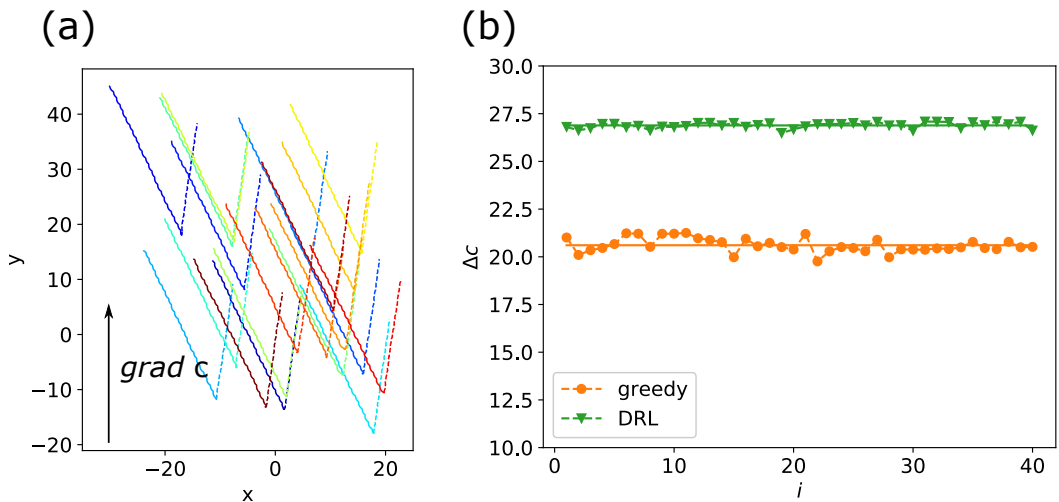}
  \caption{(a) Centrelines of some test swimming sperm cells: solid lines for the DRL; dash lines for the greedy strategy. (b) The final gains ($\Delta c=c(t_\mathrm{life})-c(0)$) of 40 sperm cells. The solid lines mark the average values. $N_T=4$, $t_\mathrm{life}=200$.}
\label{fig:sec4_trjs}
\end{figure}

Secondly, we examine the case of $N_T=4$. Again, we compare the results among the  swinging curvature pattern, the greedy strategy and the DRL. In the swinging pattern, the curvature changes every $N_T \Delta T/2$. As shown in Fig.~\ref{fig:sec4_comp} (a),   the DRL and the greedy strategy still follow the swinging pattern most of the time,
but only break the pattern occasionally. The swinging pattern still leads the sperm cell back to its original position as shown in Fig.~\ref{fig:sec4_comp} (b), whereas DRL and the greedy strategy are able to guide the sperm cell to swim toward higher concentration steadily. In the greedy strategy, the sperm cell tends to break the pattern more frequently and maintain the same curvature for a few time steps. As a consequence, the alignment of the overall migration direction of the cell to the gradient direction is better in the greedy strategy than that of the DRL,
as shown in Fig.~\ref{fig:sec4_comp} (b).
However, the overall swimming speed towards the gradient direction
turns out to be slower in the greedy strategy.
The difference in the swimming speeds or the effective swimming distances can be explained by the enlarged portion of the trajectory in Fig.~\ref{fig:sec4_comp} (b),
where the greedy strategy often leads the cell to circulate locally more than once
as indicated by the thicker orange line.
Therefore, the DRL has a significantly better performance for the gains of chemoattractant than the greedy strategy as shown in Fig.~\ref{fig:sec4_comp} (c).
Furthermore, Fig. \ref{fig:sec4_trjs} shows the centerlines of many test sperm cells and their final gains. It is apparent that sperm cells guided by the DRL always have significant higher final gains than that guided by the greedy strategy. 

\begin{figure}[hbt]
\centering
  \includegraphics[width=0.9\linewidth]{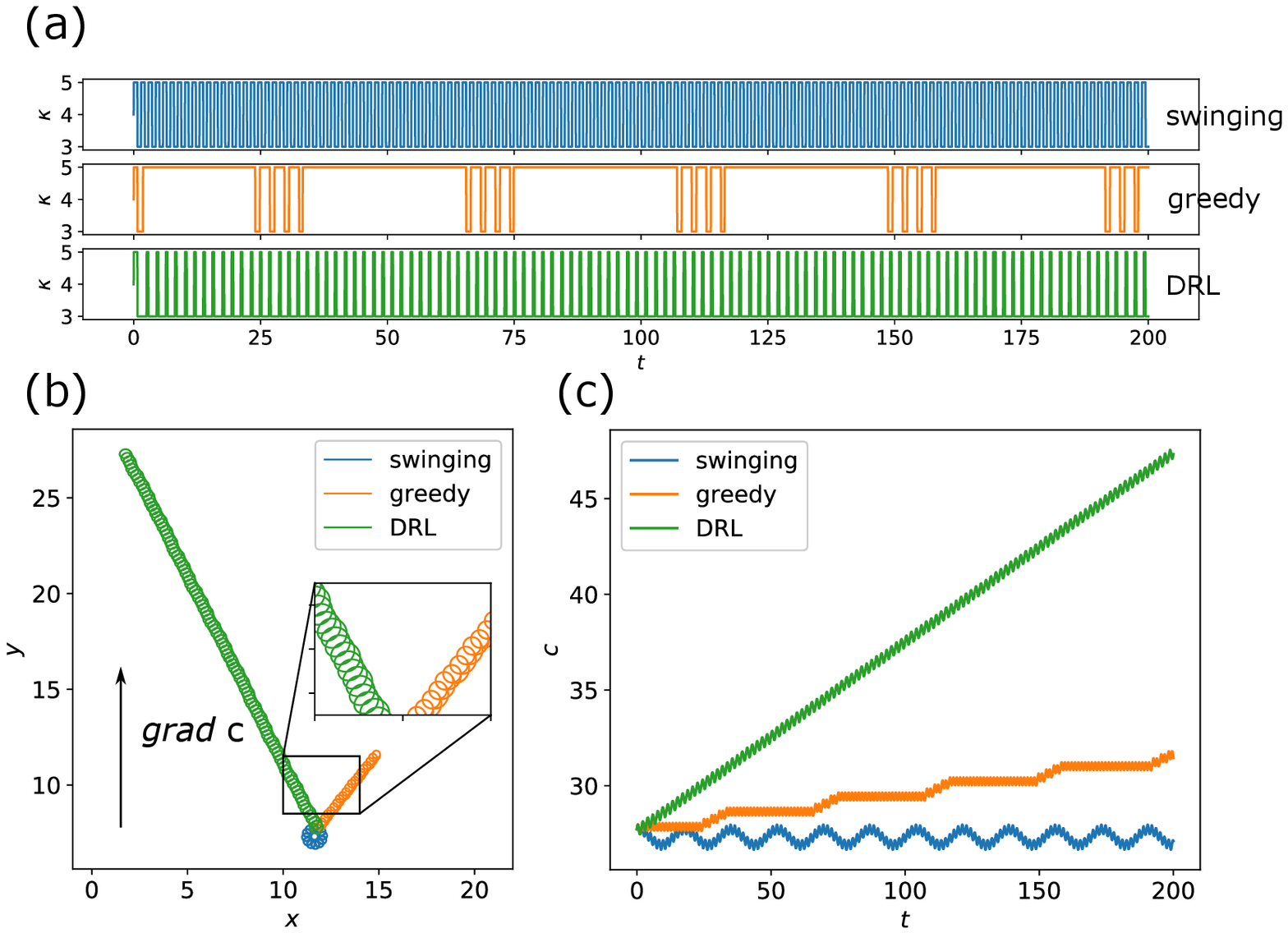}
  \caption{Comparison of (a) evolution of $\kappa$, (b) swimming trajectories, and (c) gains of chemoattractant for a sperm cell using different steering stratieges. $N_T=8$, $t_\mathrm{life}=200$.}
\label{fig:sec8_comp}
\end{figure}

\begin{figure}[hbt]
\centering
  \includegraphics[width=0.9\linewidth]{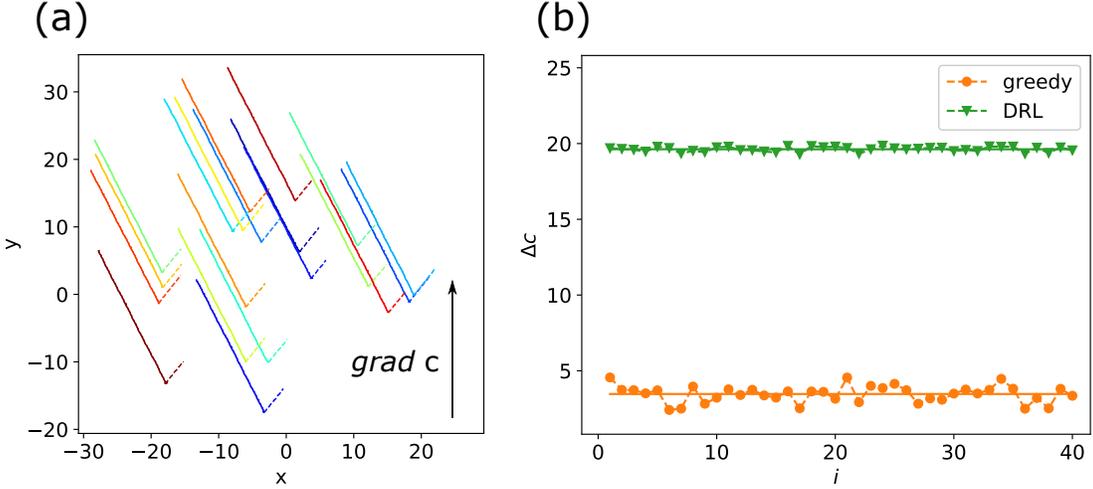}
  \caption{(a) Centerlines of some test swimming sperm cells: solid lines for the DRL; dash lines for the greedy strategy. (b) The final gains ($\Delta c=c(t_\mathrm{life})-c(0)$) of 40 sperm cells. The solid lines mark the average values. $N_T=8$, $t_\mathrm{life}=200$.}
\label{fig:sec8_trjs}
\end{figure}

Finally, we examine the case of $N_T=8$. The results are shown in Fig.~\ref{fig:sec8_comp} and Fig.~\ref{fig:sec8_trjs}. 
With more perceptions in the history, neither the greedy strategy
nor the DRL follows the swinging pattern. In this case, changing the curvature every $N_T \Delta T/2 $ is a very inefficient way to shift the path center,
as the curvature radius of the centerline is very small. Instead, both the greedy strategy and DRL follow their own patterns to change the curvature periodically, respectively. 
Again, cells guided by the DRL avoids to circulate locally and travel
longer effective distance along the gradient direction upwards.
Therefore, the DRL achives better final gains compared to that of the greedy strategy.

To understand better the three sets of results above,
let us further analyze how each strategy takes effect. 
We first note that the swinging pattern always leads to an arc/a circle for the centerline. The origin of this phenomenon is as follows. The curvature of the path is alternating every $N_T \Delta T/2$,
during which the tangential of the path turns less than $180^\circ$ for $\kappa_1$, or more than $180^\circ$ for $\kappa_2$.
This can been deduced from Eqs.(\ref{eq:T}) and (\ref{eq:Delta_T}).
%As which is not exactly 
%so every $N_T \Delta T$ the tangential of the path does not turn exactly $360^\circ$, and every $N_T \Delta %T/2$ the tangential of the path also does not turn exactly $180^\circ$.
Every time the curvature alternates, the center of the curvature deviates in the direction of the normal of the swimmer's path. Over long time of many alternating curvatures, the centers of the curvatures form its own trajectory, which is called centerline here.
The centerline of the path turns out to be an arc, or a full circle if given sufficient time.
In light of this, a sperm cell may use the following strategy to gain net migration toward the gradient direction: following the swinging pattern most of the time but breaking the pattern occasionally.
Therefore, the centerline of the path would be prevented from becoming a full circle, but becoming a curvy line formed by many broken arcs bending towards the gradient direction upwards.
This is exactly the strategy the DRL has discovered with $N_T=2$ and $N_T=4$.
We observe that the centerlines in Fig.~\ref{fig:sec2_trjs} (a) Fig.~\ref{fig:sec4_trjs} (a) are formed by connecting arcs. 
In the case of $N_T=8$, the situation is more subtle. Since the curvature radius of the centerline is very small when the swinging pattern is followed, it is inefficient to applied the strategy we just mentioned. However, as the action frequency is so high for $N_t=8$ that the DRL does not have to derive from the swinging pattern as a template. Instead, it can find a different sequence of $\kappa$ in approximately one average period of time,
which not only facilitates the sperm cell to gain net migration toward higher chemoattractant concentration but also reduces the direction change of $\mathbf{n}$ after every $N_T\Delta T$. In addition, the DRL can also perform the phase shift in a finer way due to the small $\Delta T$. These characteristics of $N_t=8$ lead to very straight centerlines in Fig.~\ref{fig:sec8_trjs} (a), whereas the centerlines are much more wavy for $N_t=2$ and $4$ on Figs.~\ref{fig:sec2_trjs}(a) and ~\ref{fig:sec4_trjs}(a), respectively.

In any case, there is an angle deviation between the overall migration direction and the gradient direction. In the case of $N_t=2$ and $4$, the centerlines are formed by arcs. The change of direction from arc to arc is a constant, therefore the arc length is the only one degree of freedom to control both the overall migration direction and the overall migration speed along the gradient direction. There is a trade-off between the overall migration direction
and effective migration speed, which leads to the angle deviation.
The swinging pattern has a good feature: the increase and decrease of the $\kappa$ are almost counterphase, so the two shift vectors in every $N_T\Delta_T$ are approximately in one line to enhance each other. 
In the case of $N_T=8$, the swinging pattern is completely discarded
so that the sequence of $\kappa$ found by the DRL may not has the nice feature of swinging pattern. This means that the net migration vectors after every $N_T\Delta T$ may be very different from the gradient direction. When the DRL tries to minimize the angle change of $\mathbf{n}$ after every $N_T\Delta T$,  hence maintaining the migration direction better, it sacrifices the migration direction to be largely deviated from the gradient direction.

In the three cases presented above we have fixed the physical simulation time $t_\mathrm{life}$ to be the same, therefore we can also compare the performance with different $N_T$. Comparing Fig.~\ref{fig:sec2_trjs}, Fig.~\ref{fig:sec4_trjs} and Fig.~\ref{fig:sec8_trjs}, we can see that increasing $N_T$ from 2 to 4 does enhance the performance of both the greedy strategy and the DRL. But the performance deteriorates if we further increase $N_T$ to 8. This indicates that smaller action interval and finer perception do not necessarily lead to better chemotaxis using the DRL or the greedy strategy.
Nevertheless, as $N_t$ increase from $2$ to $8$, the superiority of the DRL over the greedy strategy is indeed enhanced.

\subsection{Varying the swimming velocity}

\begin{figure}[hbt]
\centering
  \includegraphics[width=0.9\linewidth]{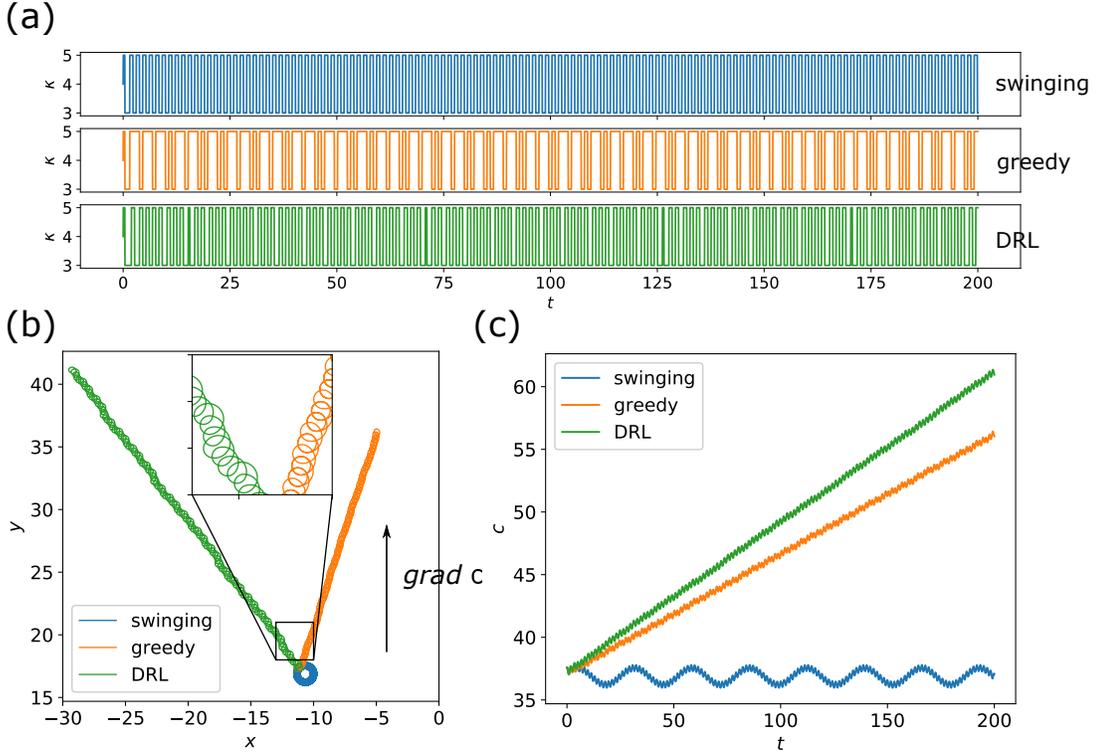}
  \caption{Comparison of (a) evolution of $\kappa$, (b) swimming trajectories, and (c) gains of chemoattractant for a sperm cell using different steering strategies. $N_T=4$ and $v_0$ is varying according to Eq.~(\ref{eq:varyingV}). $t_\mathrm{life}=200$.} 
\label{fig:sec4_varyingV_comp}
\end{figure}

\begin{figure}[hbt]
\centering
  \includegraphics[width=0.9\linewidth]{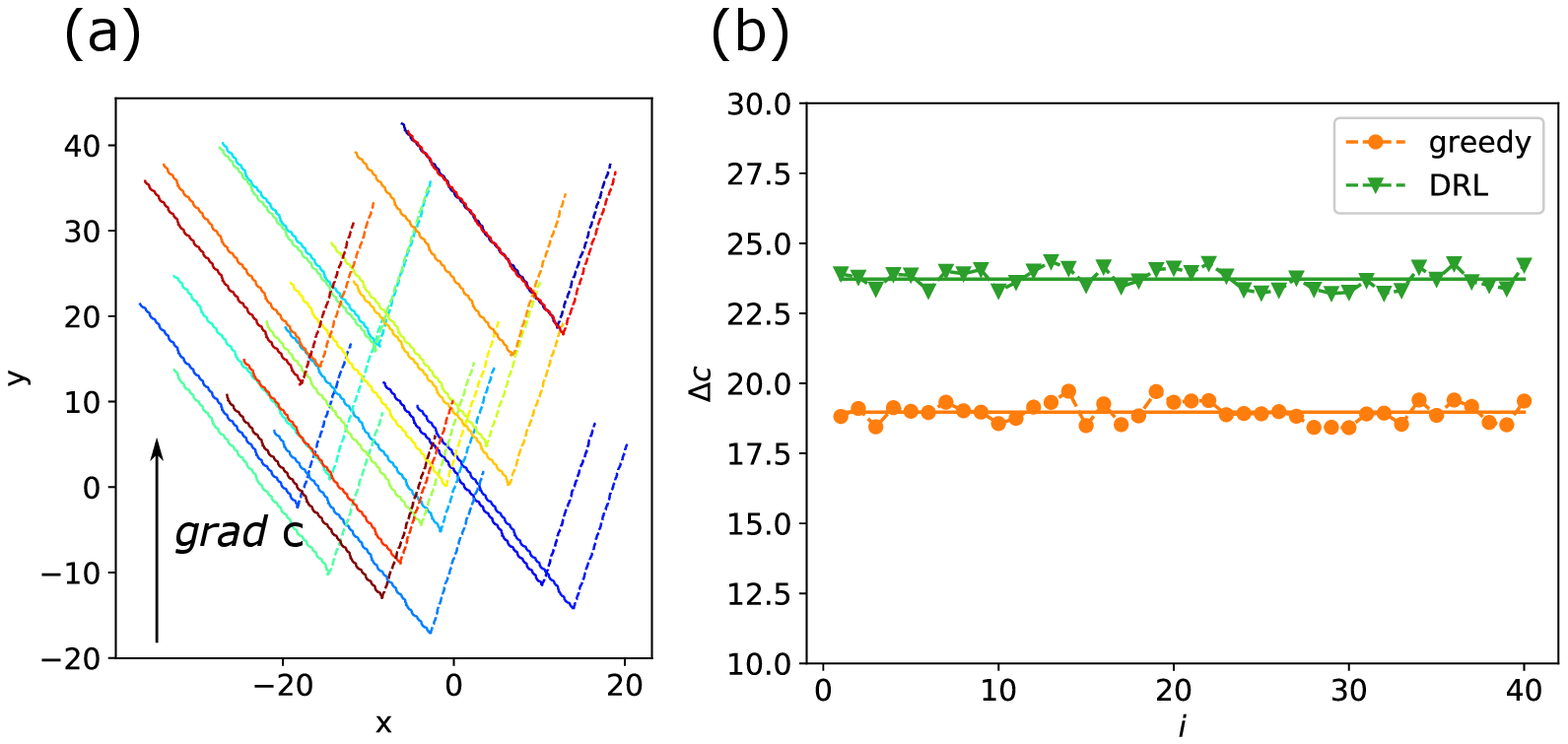}
  \caption{(a) Centerlines of some test swimming sperm cells: solid lines for the DRL; dash lines for the greedy strategy. (b) The final gains ($\Delta c=c(t_\mathrm{life})-c(0)$) of 40 sperm cells. The solid lines mark the average values. $N_T=4$ and $v_0$ is varying according to Eq.~(\ref{eq:varyingV}). $t_\mathrm{life}=200$.}
\label{fig:sec4_varyingV_trjs}
\end{figure}

Until now we have assumed that the swimming velocity $v$ is constant.
However, when a flagellum with a non-zero intrinsic curvature changes its beating waveform in reality, both its swimming velocity $v$ and the curvature $\kappa$ adapts accordingly~\cite{Liu2020POF,Liu2020PRF}. 
In line with the reality, we allow the swimming velocity to adapt as
\begin{equation}
	v = \begin{cases}
		v_1 & \text{if}\quad \kappa = \kappa_1 \\
		v_2 & \text{if}\quad \kappa = \kappa_2.
	\end{cases}
    \label{eq:varyingV}	
\end{equation}
We still set $\Delta T = \mathrm{int}(T/N_T/\Delta t)\Delta t$.
As $v$ is varying now, neither $T$ nor $\Delta T$ is constant.
Therefore, the perception/action interval varies accordingly,
which makes the problem more complex. 
We show our results in Fig.~\ref{fig:sec4_varyingV_comp} and Fig.~\ref{fig:sec4_varyingV_trjs}. It can be seen that despite the increased complexity the DRL still outperforms the greedy strategy. Comparing Fig.~\ref{fig:sec4_varyingV_comp} and Fig.~\ref{fig:sec4_comp} we can see that variation pattern of $\kappa$ has changed due to the adaptivity of the swimming speed. The overall migration direction deviates more from the gradient direction for both the greedy strategy and the DRL in the presence of adapting speed. Fig.~\ref{fig:sec4_varyingV_trjs} further indicates that the performance of both the greedy strategy and the DRL deteriorates slightly if compared with the case of constant swimming speed in Fig.~\ref{fig:sec4_trjs}.

\begin{figure}[hbt]
\centering
  \includegraphics[width=0.9\linewidth]{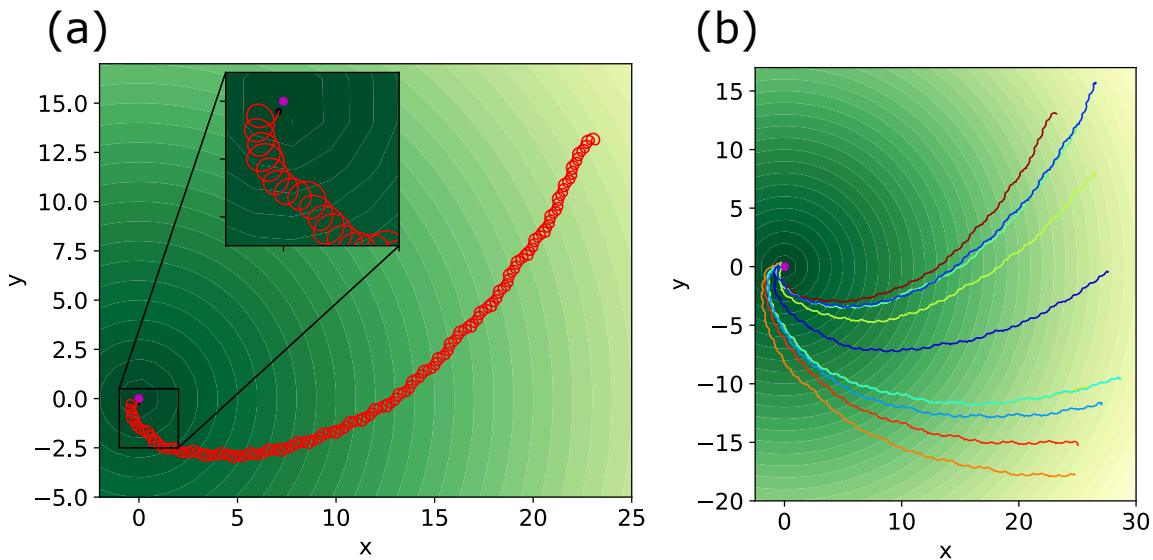}
  \caption{(a) Path of a swimming sperm cell guided by the DRL in radial concentration field (Eq.~(\ref{eq:radial_conc_field})); (b) Centerlines of the paths of 9 test sperm cells. The simulation stops only when the sperm cell is very close to the origin, hence $t_\mathrm{life}$ is not fixed here. In the figures, darker color represents higher concentration.}
\label{fig:sec4_radial}
\end{figure}

 When the chemoattractant is released from the surface of an egg cell, a radial concentration profile is usually formed. Hence we apply the Q-Network trained in the linear concentration field described by Eq.~(\ref{eq:linear_conc_field}) and test its performance in a radial one given in Eq.~(\ref{eq:radial_conc_field}). The result is presented in Fig.~\ref{fig:sec4_radial}. Figure.~\ref{fig:sec4_radial} (a) shows that the sperm cell guided by the Q-network is approaching to the egg cell and its centerline has a spiral path. The centerlines of more test sperm cells are shown in Figure.~\ref{fig:sec4_radial} (b), which carry the same spiral characteristics. This feature of spiral centerline originates from the angle deviation of the shift direction to the gradient direction. Interestingly, our DRL results bear the same feature as the chemotaxis model proposed by Friedrich and Jülicher~\cite{Friedrich2007}, which also predicts spiral centerline in 2D.


\subsection{External flow field}
\begin{figure}[hbt]
\centering
  \includegraphics[width=0.98\linewidth]{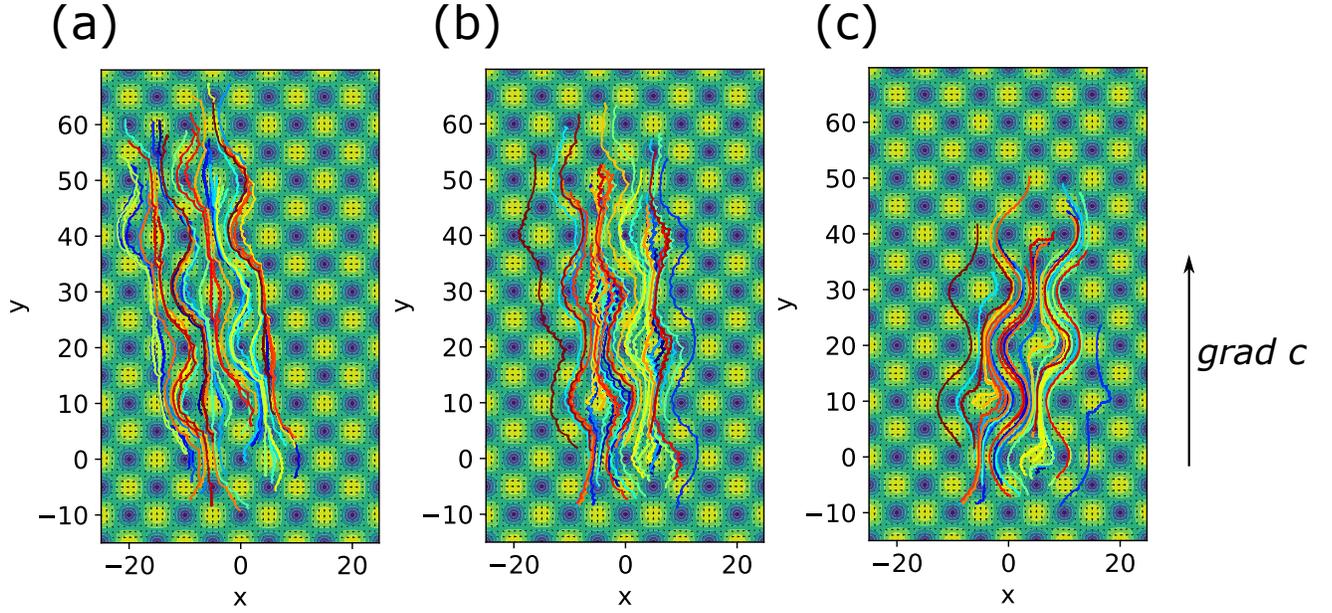}
  \caption{Centerlines of paths from 40 test swimming sperm cells in TG flow field. (a) Cells are aware of the flow field and the Q-Network is trained with the extra information of the external flow; (b) Cells are short-sighted and the Q-network is trained without information of the external flow; (c) Cells follow the greedy strategy. $t_\mathrm{life}=400$.  In the figures, brighter color represents larger external flow velocity.}
\label{fig:TG_centrelines}
\end{figure}

\begin{figure}[hbt]
\centering
  \includegraphics[width=0.5\linewidth]{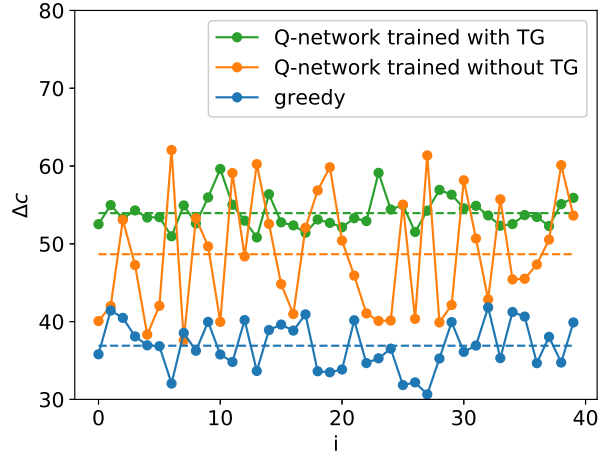}
  \caption{The final gains of the test sperm cells with the Taylor-Green (TG) external flow field present. $t_\mathrm{life}=400$.}
\label{fig:gains_TG}
\end{figure}

In the presence of an external flow field described by Eq.~(\ref{eq:taylor_green}), the swimming path of a sperm cell may be disturbed. Hence, we allow the sperm cell to sense the flow field information and examine whether the DRL can effectively utilize the external flow field to facilitate the chemotaxis. We train the Q-Network using the input of Eq.~(\ref{eq:rich_input}) for 6000 epochs and present the centerlines of swimming paths from 40 test sperm cells in Fig.~\ref{fig:TG_centrelines} (a). For comparison, the centerlines of these sperm cells guided by a ``short sighted''  Q-network are also shown in Fig.~\ref{fig:TG_centrelines} (b), where the DRL is unaware of the external flow field and take Eq.~(\ref{eq:simple_input}) as inputs. The path centrelines of these test sperm cells guided by the greedy strategy are also shown in Fig.~\ref{fig:TG_centrelines} (c). The comparison of the corresponding final gains are presented in Fig.~\ref{fig:gains_TG},
which shows that the Q-network trained with the TG has the best performance. In the case of Q-Network trained without the TG information, the flow are purely external disturbance to the swimming paths. Therefore, the TG flow may facilitate the chemotaxis sometimes and may also impede the chemotaxis at other times. As a result, the variance of the final gains is very large. When we allow the Q-Network to take into account the flow information, the swimming path avoid being impeded by the vortex. Hence the average gain is higher and the variance is significantly smaller. This indicates that the DRL is able to guide the sperm cells to actively utilize the external flow field to facilitate its chemotaxis.
Again, the greedy strategy is unsophisticated and has the poorest performance.

\section{Summary and conclusions}

We investigate how sea urchin sperm cells can self-learn chemotactic motion through the DRL alogrithm. In reality, the chemotactic motion of a sperm cell is accomplished through a very complex signalling network inside the cell. In this work we replace this biological signalling network with an artificial neural network to acts as a decision-making agent. We allow the agent to record the local chemoattractant concentrations and curvatures as states on the swimming path, and further define the increment of average concentration as reward. The agent is trained via the DRL algorithm
and therefore, the sperm cells can self-learn to accomplish chemotactic motion very similar to a realistic one: swimming in spiral paths towards higher chemoattractant concentration. Indeed, very little information is needed to accomplish this kind of chemotactic motion. If the agent utlizes the information only at the current moment and about haft an average period ago ($N_T=2$), it can learn a strategy 
similar to the greedy strategy. This strategy already enables it to actively swim towards higher chemoattractant concentration in a tortuous way. If the perception and action frequency is slightly increased ($N_T=4,8$), the sperm cell can swim in a much straighter way along the concentration gradient upwards. In any case, the DRL always discovers more efficient strategies than the greedy strategy. Nevertheless, in most cases there is a deviation angle between the net migration direction and the concentration gradient direction, which originates from the trade-off between maintaining an overall migration direction along the gradient direction and achieving an effective migration speed. This deviation angle results in a spiral centerline that circles toward the chemoattractant source in a radial concentration field. When there is an external flow field present, it disturbs the chemotactic motion and causes a large variance in the chemotactic performance. However, if we allow the agent to gather the external flow field, it can learn to actively utilize the extra information to facilitate its chemotactic motion, which increases the average chemotactic performance and decreases the variance.

Our results demonstrate that complex biological behaviours can be reproduced with an artificial neural network through the DRL algorithm with little environmental information.  It has taken millions of generations for the biological cells to adapt, evolve, and
maneuver rationally. In contrast, it takes only thousands of epochs for the DRL to develop maneuver strategies that define similar behaviours as the biological ones. Even though the implementation for the decision-making machinery in biological organisms and the DRL algorithm are completely different, the latter can still be a useful tool to reveal many possible mechanisms in biological organisms. Besides, it is also possible to use the DRL approach to design smart synthetic microswimmers that can self-learn to adjust itself to complex environments and develop ``intelligent'' behaviours. For example, elastic filament with non-zero intrinsic curvature can be used to imitate the sea urchin sperm cells. By applying different actuating frequencies, the swimming curvature of the filament can be ajusted~\cite{Liu2020POF,Liu2020PRF}. Therefore, the approach presented in this work provides a tool to use only one control variable (the actuating frequency) to guide the microswimmer to achieve complex chemotactic motions.

It should also bear in mind that the deep artificial neural network used in our model has tens of thousands neurons, while some multicellular organisms, for example, C.\ elegans, possess only a small number of neurons to achieve chemotactic motion~\cite{Itskovits2018}. If the artificial neural network were less complex, it wouldbe be more possible to be implemented in synthetic microswimmers. Therefore, the NEAT technology~\cite{Hartl2021} is perhaps pertinent to explore a simpler topology for the neural network, which will be our research direction in the near future.

\section*{Acknowledgement}
X.B. received the starting grant from 100 talents program of Zhejiang University and
also acknowledges the grant of Innovative Research Foundation of Ship General Performance 
via contract number 31422121.
\bibliographystyle{unsrt}
\bibliography{main}
\end{document}